\RequirePackage{fix-cm}
\documentclass[smallextended]{svjour3}       
\smartqed  
\usepackage{graphicx}
\usepackage{amssymb}
\usepackage{multicol}
\usepackage{algorithmic}
\usepackage{tabularx}
\usepackage{booktabs}

\usepackage{natbib}
\begin{document}

\title{A Multi-Heuristic Approach for Solving the Pre-Marshalling Problem
}


\author{Raka Jovanovic         \and
       Milan Tuba \and
        Stefan Vo{\ss} 
}


\institute{Raka Jovanovic \at
 Institute of Physics, University of Belgrade, Pregrevica 118, Zemun, Serbia\\
              \email{rakabog@yahoo.com} \\          
             \emph{Present address:} Qatar Environment and Energy Research Institute (QEERI),
  PO Box 5825, Doha, Qatar 
           \and
           Milan Tuba \at
		Faculty of Computer Science, Megatrend University, Belgrade, Serbia
		\and
         Stefan Vo{\ss} \at
         Institute of Information Systems, University of Hamburg, Von-Melle-Park 5, 20146 Hamburg, Germany \email{stefan.voss@uni-hamburg.de}
}

\date{Received: date / Accepted: date}

\maketitle

\begin{abstract}
Minimizing the number of reshuffling operations at maritime container terminals incorporates the Pre-Marshalling Problem (PMP) as an important problem.
Based on an analysis of existing solution approaches we develop new heuristics utilizing specific properties of problem instances of the PMP. We show that the heuristic performance is highly dependent on these properties. We introduce a new method that exploits a greedy heuristic of four stages, where for each of these stages several different heuristics may be applied. Instead of using randomization to improve the performance of the heuristic, we repetitively generate a number of solutions by using a combination of different heuristics for each stage. In doing so, only a small number of solutions is generated for which we intend that they do not have undesirable properties, contrary to the case when simple randomization is used. Our experiments show that such a deterministic algorithm significantly outperforms the original nondeterministic method when the quality of found solutions is observed, with a much lower number of generated solutions.
\end{abstract}

\section{Introduction}

\label{SecIntro}

In container terminals the time that is needed for loading containers to transport vehicles and vessels is of utmost importance. Terminals usually operate with a limited amount of storage space; because of this, block stacking is used to increase the space utilization.  More precisely, containers are simply stored over each other in several stacks. A problem may arise as only the top container can be retrieved from each stack. While containers usually need to be loaded to transport vehicles in a certain order, in general, not only will it be necessary to move containers from the stacks to the transport vehicles but they also have to be relocated within container bays to make retrieval in the specified order possible. The order is reflected by priorities, where a small priority value means that a container must be retrieved earlier than one with a larger priority value. The loading process may be the most effective if the number of block movements is minimized. This practical problem has been formalized in several forms like the Blocks Relocation Problem (BRP), the Re-Marshalling Problem (RMP), i.e. intra-block marshalling and the Pre-Marshalling Problem (PMP) \citep{OverView}.

The goal of the PMP is to reorder the blocks within a container bay to have all the blocks well located. We use the term ``well located" for a block if there are no blocks of smaller priority value located below it.
In this paper we focus on solving the PMP using a deterministic greedy algorithm.
In our approach we start with an algorithm described in \citep{Exposito} which uses a heuristic process for the PMP for developing a greedy algorithm, but combines it with a certain level of randomization to improve the quality of results. The developed method is similar to an approach that has previously been used for the BRP \citep{Jovanovic2014}.
%

Usually greedy algorithms give results that are of lower quality compared to more complex methods like those following the tree search or  the corridor method's paradigm. The advantage of greedy algorithms is that they, in most cases, have a much lower computational cost than other more complex approaches. The main problem with most greedy heuristics is that they only create one solution which is frequently  just a ``relatively" good solution. One possible method to avoid this problem is adding some type of randomization or even some learning process like the ant colony optimization. In the case when simple randomization is used, as in \citep{Exposito}, although an improvement is achieved, it often has the consequence that a large number of solutions is created that simply do not have good properties.

In our work we attempt to avoid the use of randomization and try to generate a large number of solutions that all satisfy some desirable properties.  More precisely, we use different heuristics in different  stages of the greedy algorithm proposed by \citep{Exposito} and in this way  a large number of solutions is generated. Heuristics in a greedy approach can loosely be defined as functions that give us good properties for choosing the next step in generating a solution. The problem is that this function just corresponds to a specific ``guess'' about what good properties are. Because of this, for many problems several competing heuristics are developed, that are suitable for different types of problem instances; see, e.g., well known approaches for the simple assembly line balancing problem \citep{SVDWsv96}. A common practice is to use several heuristics and choose the best found solution. This idea is well known, though, in the case of the PMP it can be exploited extensively as we can combine the heuristics for different stages of the algorithm.

To make this method work, special care is necessary to always find  feasible  solutions. This seems not an issue in similar greedy algorithms implemented for the BRP, but in the case of the PMP the algorithm would frequently bring the container bay into a state from which it is not possible to generate adequate solutions by simply applying the heuristic. This type of deadlock would happen in cases when there is a high level of occupancy of the bay. The standard approach to resolving this is to use a certain level of backtracking. In case backtracking is implemented the basic greedy algorithm would become very similar to the use of a tree search and loose its speed advantage. To resolve this, we add a simple look-ahead mechanism that would find feasible solutions in all cases when they are possible, from the initial state of the bay.  As this mechanism would in some cases add unnecessary relocations (or reshuffling operations), a simple correction stage is added at the end of the algorithm to improve the results.

We show in our tests that using this approach leads to less calculations than \cite{Exposito},  while achieving better results in almost all the test data sets. By adding randomization to this algorithm it might be possible to further improve the acquired results but this is not the focus of our research.

The article is organized as follows.  In the next section we give the problem formulation and a brief overview of published work. Then we provide an overview of the original heuristic given in \cite{Exposito}. In Section \ref{secmultiheuristic} we give a detailed specification of the heuristics used at different stages of our approach. This also incorporates a lookahead strategy used to guarantee creating feasible solutions, and the appropriate correction method. In Section \ref{S5} we give a comparison of the presented heuristics, and show that their combination gives significantly better results than previously published research.

\section{The Pre-Marshalling Problem}

The PMP, which we consider in this paper, is defined as follows. First we describe the problem setup  with some simplified assumptions as they are consistently used in literature \citep{Kim2006940}:
\begin{itemize}
\item All blocks (containers) are of the same size.
\item The container bay will be viewed as a two dimensional stacking area, with $W$ stacks, for which a maximal height (number of tiers) $H$ is given.
\item  The initial configuration of the container bay is known (and consists of a set $C$ of containers).
\item Only blocks from the top of a stack can be accessed.
\item Blocks can only be placed either on top of another block, or on the ground (tier 0).
\item{Each container has a priority value (which is not necessarily unique).}
\item{A container is well located if no container with a larger priority value is on top of it.}
\item A well located container can only be above other well located containers, and has a smaller priority value than all of those below it (or the same priority value as the one immediately below it).
\item The goal is to have all of the containers in the bay well located. (For the final bay layout this means that containers can be retrieved according to increasing priority values without any further relocations.)

\end{itemize}



The problem is to minimize the number of moves needed to create a container bay with only well located containers.

It has been shown that this problem is NP-hard \citep{OverView}.
In an early paper \citep{Kim1997} influencing this area of research, various stack configurations and their influence on the expected
number of rehandles are investigated in a scenario of loading import containers onto outside
trucks with a single transfer crane. For easy estimation regression equations are
proposed. 

{\sloppy There are quite a few papers proposing solution approaches for solving the PMP. This incorporates using a tree search algorithm \citep{Bortfeldt2012531}, integer programming \citep{MultiBay} or the corridor method para\-digm \citep{Coridor}. Algorithms with direct heuristics have been developed by \citep{Huang}; a neighborhood search heuristic can be found in \citep{Lee2009468}. Another heuristic is the one by \citep{Exposito}. Moreover, this paper also incorporates a simple A*-algorithm which was lateron improved and appended by some symmetry breaking rules by \citep{PacinoTierneyVoss}. Some comments on logical observations leading to a lower bound are provided in \citep{Voss2012Misoverlay}. Some of these ideas are also incorporated in the tree search algorithm of \citep{Bortfeldt2012531}.
A constraint programming approach together with a more general problem description allowing for priority ranges rather than priority values has been proposed by \citep{rendlprand}.
Note that the PMP is also closely related to blocks world planning; see, e.g., \citep{gupta:nau:92}. A more comprehensive survey on the PMP and related problems is provided in \citep{OverView} and more recently by \citep{Lehnfeld2014}.
}

\section{The Basic Heuristic Scheme}\label{secheuristic}

In this section we describe and extend the heuristic of \cite{Exposito}. The general idea of this algorithm is to well locate containers one by one, starting with the containers with largest priority value, say $p$. Note that according to the problem definition a container with a largest priority value cannot be on top of a container with a smaller priority value as it would otherwise hinder this container from being well located. The following pseudo-code gives an outline of the method.

\begin{algorithmic}
\STATE{$i=p$}
\STATE{$A_i$=Set of non well located containers with priority value $i$}
\WHILE{($i \neq 0$) }
\WHILE{($A_i \neq \varnothing$) }
  \STATE Select container $c\in A_i$
  \STATE Select a destination stack $d$ for $c$
  \STATE Well locate container $c$ from its current position to stack $d$
  \STATE $A_i = A_i \setminus \{c\}$
\ENDWHILE
  \STATE Fill destination stack;
  \STATE $i  =  i-1$
\ENDWHILE
\STATE
\end{algorithmic}

This type of algorithm can be divided into four stages which will be detailed in the following subsections.

\begin{enumerate}
\item Select a container to be well located.
\item Select a destination stack.
\item Relocate the necessary containers to make the well locating possible.
\item Filling.
\end{enumerate}

Compared to the original algorithm we introduce a new heuristic where we select which container will be well located, without the priority constraint. In the case of the original work presented in \citep{Exposito}, the selection of the next target container has been done using a random selection between the blocks with the maximal priority value.  As it will be explained below this is just a simple lookahead mechanism. In the following subsections we shall analyze the heuristics used at each stage of the algorithm.

\subsection{Selecting a Destination Stack}
\label{SDS}

\begin{figure*}[htcb]
\centering
\includegraphics[width=0.7\textwidth]{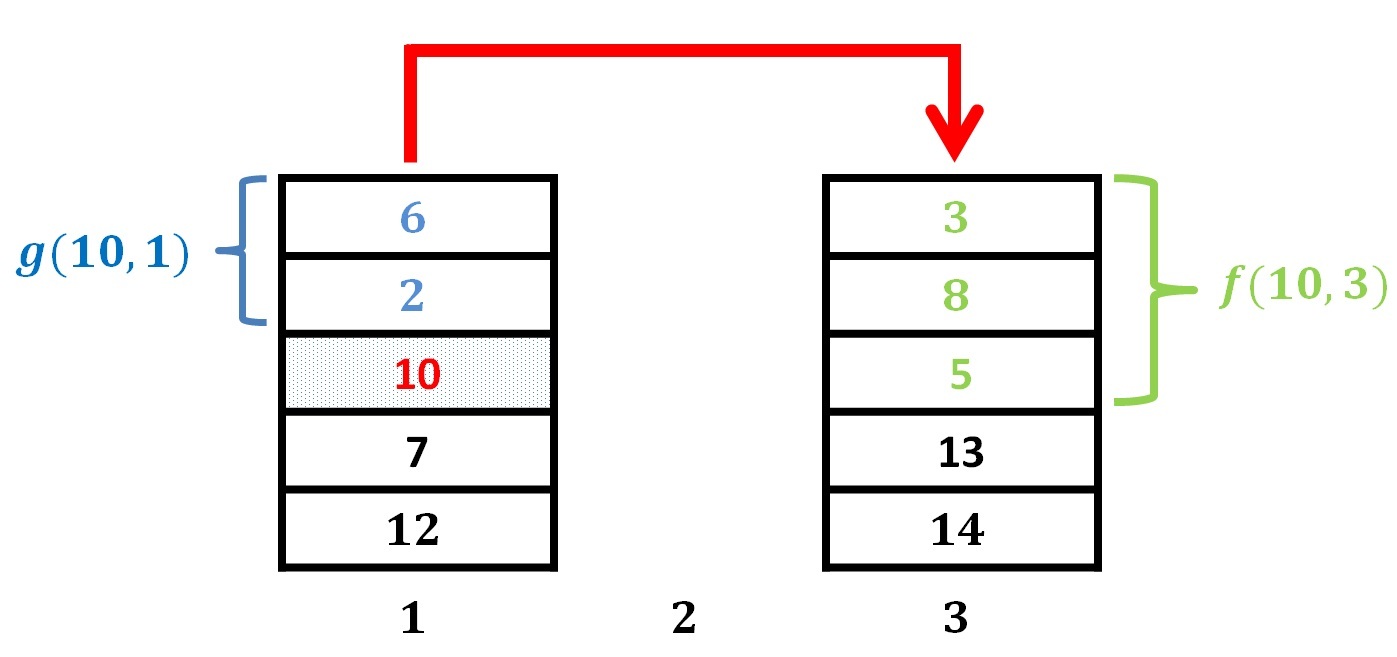}
\caption{Graphic presentation of the basic functions $f(10,3) =3$ and $g(10,1)=2$.}
\label{fig:BasicFunc}
\end{figure*}

To ease exposition, we first provide some details of how a destination stack $s^*$ is selected for block $c$ that we wish to well locate. The goal in this stage is to well locate $c$ in the smallest number of container relocations. The number of relocations is depending on two factors. First the number of containers necessary to be relocated to access block $c$, more precisely the blocks above $c$ have to be removed. Define $g(c,s)$ as the number of blocks above $c$ in stack $s$. The second factor is how many relocations we need to well locate block $c$ at some stack $s^*$. In this way we define functions $f(c,s^*)$. Practically, this is the number of blocks that need to be removed from $s^*$, to have a well located block $a$ with a larger or the same priority value than $c$ to allow $c$ to be retrieved from the final bay layout before $a$. An empty stack has the smallest such number as every block can be well located once it is put onto the ground. We give a graphic representation of functions $f$ and $g$ in Figure \ref{fig:BasicFunc}. As presented in \citep{Exposito}, a heuristic function $w$ can be presented in the following form.
\begin{equation}
\label{ChoiceExposito}
w(c,s^*)  = \left\{ \begin{array}{l l}
f(c,s^*)+g(c,s)+1 & s\neq s^* \\
f(c,s^*)+1      & s=s^*
\end{array} \right.
\end{equation}

We minimize the heuristic function $w(c,s^*)$ to determine the stack $s^*$ to which block $c$ will be well located. This will simply be the stack $s^*$ that has the minimal value of $w(c,s^*)$. It has been shown that this approach gives results of good quality \citep{Exposito}.

\begin{figure*}[htcb]
\centering
\includegraphics[width=0.6\textwidth]{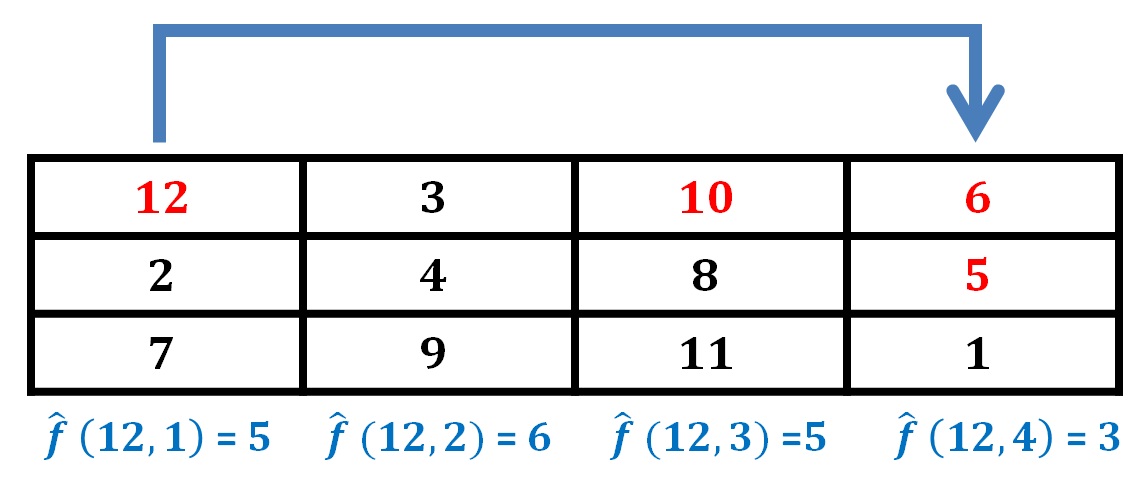}
\caption{Exemplifying the heuristic function. The values are shown for the case when block $12$ is being well located.}
\label{fig:ImprovedStack}
\end{figure*}

The problem with the heuristic function given in Eq. (\ref{ChoiceExposito}), is that it does not consider that this move can create some new, in many cases avoidable relocations. The most obvious source of this is the relocation of already well located containers. This can be illustrated in Figure \ref{fig:ImprovedStack}. When using the original heuristic  to well locate block $c$ with priority $12$ all stacks are equal, since in all the cases $f(c,s^*)=3$. But it is evident that selecting stack $2$ might be a very bad choice because three well located containers will be moved.

We introduce a new approach that takes this into account.  We first define   $nw(c,s^*)$ as a number of well located containers that need to be relocated when $c$ is moved to stack $s^*$:

\begin{equation}
\label{NonWellLocated}
  nw(c,s^*)  = \left\{ \begin{array}{l l}
0, & \,\,\,\,\,\ ,f(c,s^*)< nwl(s^*)\\
f(c,s^*) - nwl(s^*)      & \,\,\,\,\,\ ,otherwise
  \end{array} \right.
\end{equation}
In Eq. (\ref{NonWellLocated}), $nwl(s^*)$ denotes the number of non-well located blocks in stack $s^*$.
Using the  $nw(c,s^*)$ we define a new heuristic for the number of relocations related to well locating $c$ in stack $s^*$. 
\begin{equation}
\hat{f}(c,s^*) = f(c,s^*) + nw(c,s^*)
\end{equation}
Note that it is not necessary to use a similar extension of function $g$ as we know there are no well located blocks above block $c$. The improved heuristic $\hat{w}(c,s^*)$ is defined by substituting $f$ by $\hat{f}$ in $w$. Using the improved heuristic, we can differentiate between the stacks in Figure  \ref{fig:ImprovedStack}, and choose stack $4$ as it has the smallest value of $\hat{w}(c,s^*)$.

\subsection{Selecting the Block to be Well Located }

In our approach we introduce the  use of a heuristic function for selecting which block will be well located next. This is an adaptation of the original algorithm in the sense that we remove the constraint that we only select a block $c$ that has the highest priority value. In \citep{Exposito}, the blocks are well located in descending order of their priority values. The idea of this approach is that once a container with the highest priority value is well located it will no longer interfere with the well locating of succeeding blocks, and there will be a lower number of forced relocations. Although this approach proves to be very efficient, this rule can be considered overly strict.

The reasoning for the new stage of the algorithm is the following. In many cases well locating the  container with the highest priority value can be hard in the sense that many relocations need to be performed. It may be advantageous to first well locate some other block, since the relocations that will be performed may in some cases make it easier for well locating a container with highest priority value.

\begin{figure*}[htcb]
\centering
\includegraphics[width=0.9\textwidth]{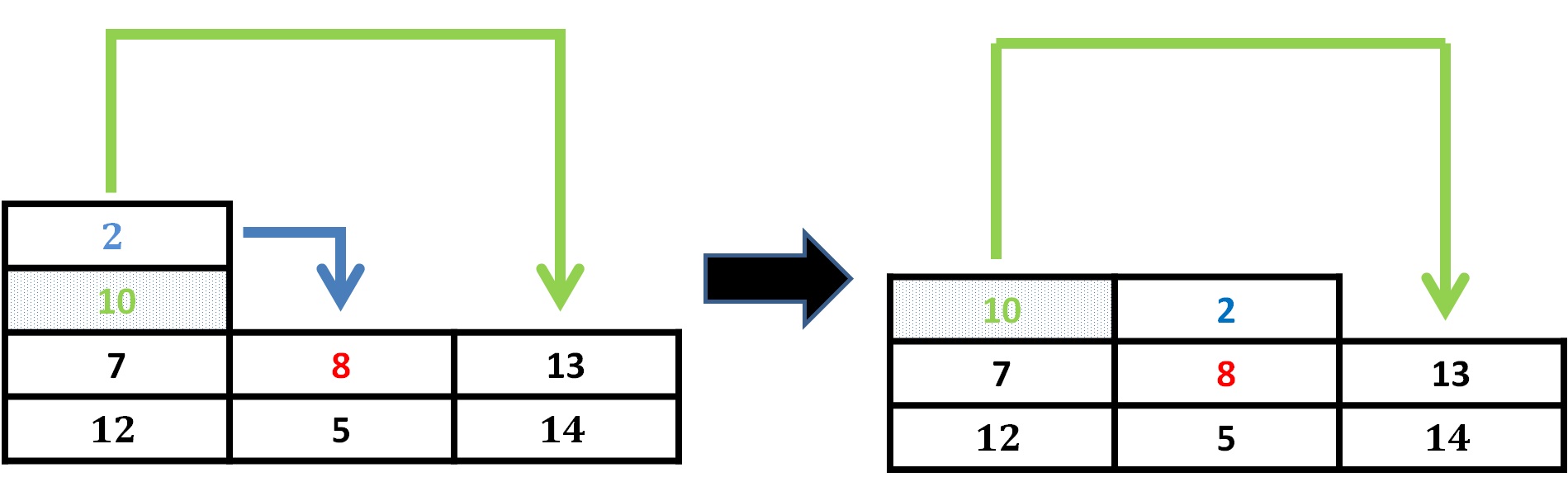}
\caption{Illustration of creating a forced relocation. When trying to well locate block $10$ on stack $3$, block $2$ needs to be positioned over block $8$ which is non well located. The new position of $2$ creates a forced relocation since block $2$ will be moved again when well locating block $8$. }
\label{fig:ForcedRelocation}
\end{figure*}

Another aspect that should be considered when well locating a block from stack $s$ to stack $s^*$ is how many new forced relocation will be created. We shall consider that we have created a forced relocation if we move block $a$ over a non-well located block $b$, and the priority value of $a$ is smaller than the one of $b$. This is due to the fact, that $b$ will be well located before $a$, and while  conducting the necessary reshuffle operation to well locate $b$, block $a$ will  be relocated. We give an example of a forced relocation in Figure \ref{fig:ForcedRelocation}.

In this section we introduce a new heuristic function that addresses this problem. The new heuristic function needs to have the following properties:
\begin{itemize}
\item Prefer well locating containers with high priority values.
\item Prefer a low number of relocations.
\item Prefer a low number of forced relocations.
\end{itemize}

The proposed heuristic can be formalized by the following function:
\begin{eqnarray}
    \label{laBlock}
    d = ind(min_{s\in S}\hat{w}(c,s))\\
    \label{laBlock2}
    \hat{h}(c)=  -p(c)+\hat{w}(c,d)+fr(c,d)
  \end{eqnarray}

In Eq. (\ref{laBlock}), $d$ is the index of the stack that has the lowest value $\hat{w}(c,s)$ for a block $c$, or in other words the stack to which block $c$ can be well located with a minimal number of relocations. Eq. (\ref{laBlock2}) represents the heuristic function for selecting the block that will be well located next. In Eq. (\ref{laBlock2}), $fr(c,d)$ gives the number of created forced relocations corresponding to the selection of stack $d$. As illustrated in Figure \ref{fig:ForcedRelocation}, $fr(10,3) = 1$. $p(c)$ is the priority value of block $c$, the negative prefix is used since we wish to minimize our heuristic function and high propriety values are more desirable.  Finally, the block that will be well located next is the one that has the minimal value of  $\hat{h}(c)$, as given in the following equation.
\begin{equation}
  \label{laMin}
    next=  ind(min_{c\in C}(\hat{h}(c))
  \end{equation}

Forced relocations have been defined in a relatively simple way in the attempt to make the calculation of $fr(c,d)$ straightforward.  A good approximation of the  total number of relocations can be effectively calculated only by tracking the priority value at the top of each stack. If a block $c$ is relocated a forced relocation occurs only if all of the top stack priority values are higher than $p(c)$. If all the blocks in a stack are well located we will consider that stack having priority value zero.

Note that the minimization  in Eq. (\ref{laMin}) can in most cases be calculated by evaluating $\hat{h}(c)$ for a small number of blocks. More precisely the highest number of  blocks that are tested is

\begin{equation}
\label{MaxTest}
 n=\hat{w}(c_m,d)+fr(c_m,d).
 \end{equation}

 In Eq. \ref{MaxTest}, $c_m$ is used for the block with the highest priority value. As it can be seen in Eq. \ref{laBlock2} the heuristic function $\hat{h}$ is dependent on the priority $p(c)$ of the block being relocated. It is obvious that for block $a$ which has a priority $p(a)< p(c_m) -n$, even if  $\hat{w}(a,d)+fr(a,d)=0$,  we have $\hat{h}(a)>\hat{h}(c_m)$.

\subsection{Relocating the Necessary Blocks}

The heuristic functions presented in the previous subsections tell us  that block $c$ should be well located in stack $s^*$. To perform this action it is necessary to relocate several containers from the source $s$, where block $c$ is located, and destination stack $s^*$ as explained in Subsection~\ref{SDS}. The goal is to relocate all the required blocks without creating new avoidable relocations in the future. This process can be divided into two parts.

\begin{itemize}
\item{Order in which blocks are relocated.}
\item{Selection of the stack to which a block will be relocated.}
\end{itemize}

In \citep{Exposito} a detailed description of the ordering in which the blocks are relocated is presented. The basic idea is that at each step we relocate one of the two top blocks of stacks $s$ or $s^*$ that has a higher priority value. Of course, only the blocks whose relocation is necessary for well locating $c$ are considered. The process is continued until all the required blocks are relocated. Using this approach the number of blocks to be moved in future iterations is minimized.

The second part is about deciding whereto blocks should be relocated. This stack is selected by some heuristic function that measures their desirability, in the sense that we do not wish to create new unnecessary relocations. This problem is very similar to what appears in the BRP. Heuristic functions of this type have been widely researched and analyzed for this problem. As a consequence we can use these heuristics in the case of the PMP.  Several different heuristics have been developed for which detailed descriptions can be found in literature.  We give a short overview of the ones that seem most suitable for the PMP.

\begin{itemize}
\item{{\it The Lowest Position (TLP) heuristic} \citep{HeuristicTLP}. In the TLP we relocate the block to a stack that has the lowest number of tiers. The goal is to keep the container bay as balanced as possible. In this way the average number of relocations should stay low, and to  avoid extreme cases where a large number of blocks needs to be moved from a stack with many tiers.}
\item{{\it Lowest Priority Index  (LPI) heuristic} \citep{Exposito}.  In this approach, the blocking block will be moved to the stack in which it blocks the highest priority value of a non well located  block.  It is expected that the overall number of reshuffles will be lowered since every time a block is put over another with a lower priority value, extra reshuffles need to be done  \citep{BeamSearch}.}
\item{{\it The Min-Max heuristic} presented in \citep{Caserta1}, and a very similar approach in \citep{HeuristicTurci}, only takes into account the maximal priority value of a block in each stack. An extended version, including a look ahead mechanism, of this algorithm has been presented \citep{Jovanovic2014}. The Min-Max heuristic when adapted to the PMP, will  also use the highest priority value of a non well located  block. It has a different way of choosing the stack to which the block will be moved to, depending if it will be blocking some new container. The general idea of this approach is to avoid relocations of blocks in the near future while grouping blocks of similar priorities.}
\end{itemize}

In practical applications of these heuristics to the PMP, certain improvement can be achieved by adding some fine tuning. First, in the case when the source and destination stack are the same, i.e.\ $s=s^*$, special care should be taken when temporarily relocating the block $c$ that is being well located. In this case the stack should be taken which has the worst value of the previously defined heuristics. In the case of the second two heuristics, reaching the top tier of a stack should be avoided.

\subsection{Filling}

In \citep{Exposito} the idea of stack filling is introduced to exploit the fact that after a specific container is relocated to a well located position it is at the top of a stack and the whole stack is well located. More precisely, when a target container $c$ is relocated in a destination stack $s^*$, this container is at the top of stack $s^*$.

\begin{figure*}[tcb]
\centering
\includegraphics[width=0.9\textwidth]{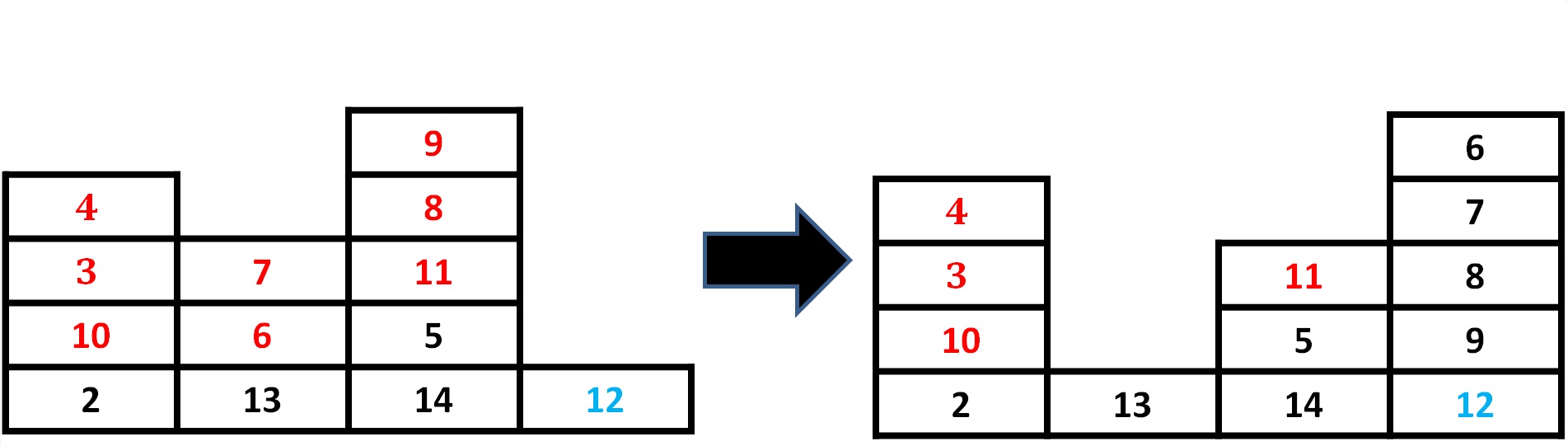}
\caption{Illustration of filling a stack. The filling is done after block $12$ has been well located.}
\label{fig:Fill}
\end{figure*}

The idea is that we can take advantage of the empty slot in a well located stack, by maximizing the number of well located containers and maximizing the number of usable slots. In practice this is equivalent to filling the empty slots in the destination stack  with non-well located containers in the most adequate sorted sequence. So for each destination stack $s$, all accessible non-located containers with a priority value equal to or lower than the container at the top of $s$ are candidates that can potentially be moved to this stack. The filling is done step by step, by relocating the container with the highest priority value that is possible to well locate to the stack $s$. The process is finished either when the stack is full or there are no more containers that can be properly located in stack $s$. An illustration of this process is given in Figure~\ref{fig:Fill}.

Tests conducted in \citep{Exposito} show that this approach is very effective, in a vast majority of problem instances, and manages to significantly reduce the number of necessary relocations needed to set the container bay in a suitable state. They have also pointed out that the use of filling degraded the quality of solutions in case of problems of small size. There are two main reasons for the possible negative effect of stack filling. First, the filling can be responsible for more block relocations that the heuristic used for well locating of specific containers explained in the previous section. This has the effect that the important heuristics presented previously can be neglected to a large extent. The other problem is that the minimal priority value of a well located stack can decrease too rapidly. This way we can lose the possibility of well locating a block $a$ with a high priority value to this stack $s$. This can in many cases result in having to add new relocations, for well locating $a$, in some case even by relocating blocks from stack $s$ that have just been placed there.

We try to balance out the positive and negative properties of stack filling by using four different types of stack filling.
\begin{itemize}
\item{\textbf{None.}} In this case no filling is conducted
\item{\textbf{Standard.}} We try to well locate as many non well located blocks to stack $s$. This is the approach as presented by \citep{Exposito}.
\item{\textbf{Safe.}} In this case we only use filling if we expect it to loose only a few slots where blocks with high priority value can be well located. We shall consider no such slots are lost if the stack $s$ has reached the maximum tier after filling. In practice we shall consider our heuristic for the desirability of performing filling, $f(s) = MaxTier - Tier(s')$, where $s'$ presents stack $s$ after filling. Filling of $s$ is only done if $f(s) \le a$, where $a$ is a predefined parameter.
\item{\textbf{Stop.}} In this case a simple lookahead mechanism is used to discontinue the filling process. More precisely, if at step $n$, we are relocating block $a$ and realize that underneath  it is $b$, and  $b>a$, and $b$ can be well located on stack $s'$.
\end{itemize}

\section{Multi-Heuristic Approach}\label{secmultiheuristic}

In the previous section we have presented several heuristics that can be used for solving the PMP. A greedy algorithm that uses one specific heuristic function for each stage of the algorithm is presented in \citep{Exposito}. The performance of the basic deterministic algorithm is improved by adding a certain level of randomization when using the heuristic function. This is done by randomly selecting one of the $n$ stacks with the highest value of the heuristic function at each step of the algorithm. In this way solutions of higher quality are found by searching a wider range of potentially good solutions. Although this method significantly improves the performance, it often explores solutions of low quality. This is due to the fact, that we often select stacks which have  undesirable properties, at least according to the heuristic function used.

In our new approach the goal is to avoid generating solutions for which we expect that they have undesirable properties. In other words, we wish only to generate solutions for which it is presumed that they will  be of good quality, while not making the original greedy algorithm more complex. For most problem instances our main focus is to find the best possible heuristic function. As it will be shown in the following section in the case of the PMP, none of the proposed heuristics is overwhelmingly superior to the competing ones. Another problem with using heuristic functions is that the use of filling greatly changes the state of the bay that we have used for evaluating the heuristic functions and as a consequence makes our choices less valid.

It is well known that if we have  several competing heuristics for some instance, their performance will be highly dependent on the specific properties of the instances we are solving. In case we do not know, or it is hard to evaluate the properties, of a problem instance, a common practice is to use several heuristics and just choose the best found solution. This simple logic can be very efficiently exploited in case of using the proposed algorithm and heuristics for the PMP.

  The idea is to test a relatively small group of good candidates for optimal solutions. In case of our problem this can be done by combining different heuristics at different stages of the algorithm. With this simple method we can generate $a*b*c*d$ different solutions that have desirable properties in different frames of reference. $a,b,c,d$ give us the number of different heuristics for each stage of the algorithm, and $a*b*c*d$ represents the total number of combinations. Using the proposed heuristics we have a total of 48=$2*3*2*4$ generated solutions.

For such a method to work it is necessary for all runs of the algorithm to create feasible solutions. In the case of BRP, when a problem instance has been well defined, a feasible solution is always acquired by a greedy algorithm using some of the heuristics proposed in \citep{HeuristicTLP,Murty,HeuristicTurci,Caserta1}. Contrary to this in case of the PMP, in many cases it is not possible to well locate all the blocks by directly applying  the greedy algorithm. This is especially noticeable for bays with a high level of occupancy.  The standard approach  for avoiding this situation is to use some kind of backtracking. There are two main drawbacks of this approach. First, the calculation time can in some cases become very long and is in general unpredictable. On the other hand when backtracking is added to the original greedy algorithm it becomes more complex. In this situation it may just be better using a more complicated method like tree search that generally gives better results.

From this we can see that by adding backtracking we have lost the two main advantages of the greedy algorithm, i.e.\ its speed and simplicity of implementation. We can avoid such deadlock using a much simpler logic. We can add a simple mechanism that can always make one more relocation possible from the source stack. In this way we can always bring the bay to a state where all the blocks are well located. One of the down sides of such approach is that often we will add several needless relocations to the final solution. The generated solutions can then be improved by adding a simple correction method.

\subsection*{Avoiding Deadlocks}

For a well defined  problem instance, in a sense that a feasible solution exists, a deadlock, i.e.\ a state of the bay from which it is not possible to add a move using a heuristic function, is only possible if the source stack $s$ is equal to the destination stack $d$ when trying to well locate block $c$. The reason for this is that when performing the necessary relocations some free slots in the bay can be lost when block $c$ is temporarily relocated. The reason for this is that we will not be placing any new block $a$ above $c$, and as a consequence $MaxTier - t(c)$ will be lost. Here $t(c)$ gives us the tier of block $c$.  Because of this if there is a stack with only one free slot block $c$ should always be relocated there.

If such a stack did not exist, when $c$ was relocated it is possible to enter a deadlock, a state from which it is not possible to perform a desired move. This situation can be avoided by a few simple steps. First we can remove the last relocation from the solution. Let us say that the removed relocation was $(s,s_r)$. After reverting the last step we know that stack $s_r$. We can select a random full stack $s_f$, and relocate a block from it to $s_r$. Now we can relocate block $c$ to this stack. In this way a new free slot has been created. This process can be more formally presented in the following way.
\begin{eqnarray}
\label{TLP}
    Except =  \{s, s(c))\}\\
    s_f      = Random(Full(  S \backslash Except))
\end{eqnarray}
And we add the following reshuffle operations to the solution.
\begin{equation}
  (s_f, s(c_{r})),  (s(c), s_f), (s, s(c))
\end{equation}

\subsection*{Correction}

Due to the fact, that the previously described mechanism would in some cases add unnecessary relocations to the solution a simple correction stage is added at the end of the algorithm to improve results.
\begin{eqnarray}
  (a,b),(b,c)\rightarrow (a,c)\\
  (a,b),(b,a)\rightarrow \varnothing
\end{eqnarray}







\section{Experimental Results} \label{S5}

All of the algorithms have been implemented in C\# using Microsoft Visual Studio 2012. The calculations have been done on a machine with Intel(R) Core(TM) i7-2630 QM CPU \@ 2.00 Ghz, 4GB of DDR3-1333 RAM, running on Microsoft Windows 7 Home Premium 64-bit.

The test data sets are the same as the ones used in \citep{Exposito}, more precisely the data sets that had unique priority values.\footnote{Note that the data from that paper had been lost and replaced by those on the webpage of the authors: https://sites.google.com/site/gciports/premarshalling-problem/bay-generator}
Tests have been conducted for a wide range of bay sizes with different proportions of maximal tier and number of stacks. For each of the bay sizes there are 40 different problem instances, and the average number of reshuffle operations is observed. In our experiments we analyze the behavior of different heuristics for each stage of the algorithm. In the final group of our tests we compare the results presented in \citep{Exposito} to the multi-heuristic approach. The calculation time for all the heuristics is very similar and very fast.

We first observe the effect of using different heuristics for the relocating of necessary blocks to perform the well-locating of some block. In these tests, to give a better evaluation, no filling or heuristic selection of blocks is used. The improvement is not used for the selection of the destination stack. The results can be seen in Table \ref{table:NecReloc}.

\begin{table}[htb]
\footnotesize
\center
\caption{\label{table:NecReloc}Comparison of heuristics for the relocating of necessary blocks when attempting to well locate a container. TLP corresponds to the Lowest Position (TLP) heuristic \citep{HeuristicTLP}, LPI represents the Lowest Priority Index  heuristic  \citep{Exposito} and MinMax is used for the heuristic presented in  \citep{Caserta1, HeuristicTurci}}
\begin{tabularx}{338pt }{X X X X X }
\toprule
Tier *Stack&    MaxHeight    &    TLP    &LPI&    MinMax\\
\midrule
3*3    &5&    11.98&    11.85&    \textbf{10.95}\\
3*4    &5&    12.20&    11.80&    \textbf{11.63}\\
3*5    &5&    13.75&    12.80&    \textbf{12.55}\\
3*6    &5&    15.70&    14.48&    \textbf{14.13}\\
3*7    &5&    17.83&    16.43&    \textbf{15.98}\\
3*8    &5&    19.63&    16.98&    \textbf{16.65}\\
4*4    &6&    23.35&    23.25&    \textbf{22.90}\\
4*5    &6&    29.15&    27.45&    \textbf{26.30}\\
4*6    &6&    30.68&    27.93&    \textbf{27.10}\\
4*7    &6&    35.50&    31.08&    \textbf{30.25}\\
5*5    &7&    45.20&    42.93&    \textbf{41.88}\\
5*6    &7&    56.35&    51.78&    \textbf{50.38}\\
5*7    &7&    61.03&    51.90&    \textbf{49.93}\\
5*8    &7&    69.23&    60.68&    \textbf{58.33}\\
5*9    &7&    75.20&    64.43&    \textbf{61.45}\\
5*10&7&    81.80&    69.23&    \textbf{65.68}\\
6*6    &8&    84.15&    77.03&    \textbf{74.98}\\
6*10&8&    123.08&    102.75&    \textbf{96.33}\\

\bottomrule
\end{tabularx}
\end{table}
First noticeable issue for the results given in  Table \ref{table:NecReloc} is that the MinMax heuristic manages to outperform the other two methods for all the bay sizes, when the average number of reshuffle operations is observed. The advantage of using the MinMax heuristic is more significant in larger problem instances. We wish to point out that when the results are observed for individual problem instances, the other two methods have achieved the best results in several of them but had very bad performance in others when compared to MinMax.

In the second group of tests we compare the effect of the two improvements for heuristics presented in \citep{Exposito}. More precisely, we observe the
influence of using a lookahead for selecting which block will be well located next and including the number of moved well located blocks when selecting the destination stack. The results are presented in Table \ref{table:Look}.
\begin{table}[htb]
\footnotesize
\center
\caption{\label{table:Look} Comparison of the effect of improvements to the  MinMax heuristic. In the notation  letters L, W are used to specify  if some improvement is added: L if a look ahead is included; W is included if the number of relocated well located containers is considered.}
\begin{tabularx}{338pt }{X X X X X X}
\toprule
Tier *Stack&    MaxTier &MinMax    &    MinMax-W    &MinMax-L&    MinMax-LW\\
\midrule
3*3    &5&    10.95&    10.95&    \textbf{10.75}&    \textbf{10.75}\\
3*4    &5&    11.63&    11.23&    11.40&    \textbf{11.08}\\
3*5    &5&    12.55&    \textbf{12.20}&    12.53&    \textbf{12.20}\\
3*6    &5&    14.13&    13.98&    14.00&    \textbf{13.75}\\
3*7    &5&    15.98&    \textbf{15.28}&    15.83&    15.50\\
3*8    &5&    16.65&    \textbf{16.25}&    16.73&    16.28\\
4*4    &6&    22.90&    22.70&    21.30&    \textbf{20.85}\\
4*5    &6&    26.30&    25.88&    24.90&    \textbf{24.63}\\
4*6    &6&    27.10&    26.48&    26.90&    \textbf{25.85}\\
4*7    &6&    30.25&    30.43&    \textbf{29.48}&    29.55\\
5*5    &7&    41.88&    41.55&    36.25&    \textbf{35.98}\\
5*6    &7&    50.33&    49.63&    43.15&    \textbf{42.35}\\
5*7    &7&    49.93&    47.85&    45.98&    \textbf{45.50}\\
5*8    &7&    58.33&    57.45&    52.68&    \textbf{52.60}\\
5*9    &7&    61.45&    61.30&    57.23&    \textbf{56.83}\\
5*10&7&    65.68&    64.98&    60.93&    \textbf{60.55}\\
6*6    &8&    74.98&    73.60&    60.70&    \textbf{59.88}\\
6*10&8&    96.33&    95.20&    84.70&    \textbf{83.38}\\

\bottomrule
\end{tabularx}
\end{table}

Table \ref{table:Look} represents the effect of each of the improvements separately, and combined when the MinMax heuristic is used. In it  MinMax represents the basic algorithm, MinMax-W is used if  we take into account the number of well located containers, MinMax-L if we select which block will be well located next and MinMax-LW if both improvements are applied. The results show that a MinMax-LW gives the best results. It is also shown that each of the improvements manage to reduce the total number of relocation operations when compared to the basic method. MinMax-W is more effective in case of smaller problem instances, while MinMax-L gives the highest level of improvement in case of large problem instances. The use of MinMax-L manages to decrease the number of reshuffle operations close to 20\% in the case of the largest problem instances. Note that although MinMax-LW gives the best results it is not overwhelming and in 25\% of the tests it is better using only one of the improvements.

In Table \ref{table:Fill} we give results of our experiments regarding different methods of filling. We also investigate the effect of combining with the lookahead mechanism. In all of the tests we use MinMax-W as the base heuristic for the algorithm.

\begin{table}[htb]
\footnotesize
\center
\caption{\label{table:Fill}Evaluation of the effect of different filling algorithms and their combination with a look ahead mechanism. In the notation an added letter L means a look ahead is included. }
\begin{tabularx}{338pt }{X X X X X X X X}
\toprule
Tier *Stack&    MaxHeight    &    None    &Standard&Stop&    Safe&    L-Stop & L-Safe\\
\midrule
3*3    &5&    10.75&    10.98&    10.80&    10.80&    10.78&    \textbf{10.68}\\
3*4    &5& 11.23&    11.75&    11.78&    11.33&    11.25&    \textbf{11.08}\\
3*5    &5&    \textbf{12.20}&    12.53&    12.55&    12.38&    12.70&    12.63\\
3*6    &5&    13.98&    14.08&    14.03&    \textbf{13.90}&    13.98&    14.03\\
3*7    &5&    \textbf{15.28}&    16.18&    16.18&    16.00&    16.28&    16.08\\
3*8    &5&    \textbf{16.25}&    16.50&    16.48&    16.50&    16.85&    16.63\\
4*4    &6&    22.70&    20.78&    20.78&    \textbf{20.60}&    21.03&    20.83\\
4*5    &6&    25.88&    24.60&    24.65&    24.38&    \textbf{23.90}&    24.13\\
4*6    &6&    26.48&    25.75&    25.73&    25.98&    \textbf{25.35}&    25.60\\
4*7    &6&    30.43&    29.70&    29.83&    29.90&    29.60&    \textbf{29.45}\\
5*5    &7&    41.55&    34.33&    34.35&    35.35&    34.40&    \textbf{34.23}\\
5*6    &7&    49.63&    40.75&    40.78&    42.53&    \textbf{39.15}&    41.18\\
5*7    &7&    47.85&    43.33&    43.28&    44.80&    44.53&\textbf{42.75}\\
5*8    &7&    57.45&    \textbf{49.20}&    49.28&    51.50&    49.48&    51.03\\
5*9    &7&    61.30&    53.30&    53.15&    55.28&    54.78&\textbf{52.90}\\
5*10&7&    64.98&    57.20&    57.35&    58.98&    58.13&\textbf{57.00}\\
6*6    &8&    73.60&    53.20&    \textbf{53.10}&    59.75&    53.28&    55.83\\
6*10&8&    95.20&    76.03&    76.05&    80.63&\textbf{75.73}&    78.73\\
\bottomrule
\end{tabularx}
\end{table}

Results acquired by using different methods of filling are much less conclusive than the ones presented in the previous two tables. First, we have confirmed the results of \citep{Exposito} that in case of small problem instances it is often advantageous not to use filling, and that it is very effective in case of large instances. From our observations of individual problem instances for problems of this size it is noticeable that the use of filling, in certain situations, would decrease the available number of free slots in the bay which had as a consequence the increase of the number of reshuffle operations needed to bring the bay to a desired state. This problem could be, to a large extent, avoided by using the lookahead mechanism. Although the results of using different filling methods would be very dependent on problem instances, combining it with lookahead is generally a good approach.

Finally, a comparison of the multi-heuristic approach and the original methods presented in \citep{Exposito} is given in Table \ref{table:Multi}. In this table the notation "Exposito D" has been used for our deterministic implementation of the algorithm given by \citep{Exposito}. The algorithm is implemented as a simple greedy algorithm excluding the randomization that is used in the original work.  It is important to point out that the results for our method are obtained generating only 48 different solutions compared to the 150 in case of \citep{Exposito}. The average results obtained using the multi-heuristic approach are noticeably better, in many cases close to 10\% improvement, than the previously published work. These results are also significantly better than the ones acquired by any of the individual heuristics. This confirms the high level of dependance between the performance of a heuristic and the initial state of the bay.

\begin{table}[htb]
\footnotesize
\center
\caption{\label{table:Multi}Comparison of the proposed multi-heuristic approach with previously published results.}
\begin{tabularx}{338pt }{X X X X X X X}
\toprule
Tier *Stack&    MaxHeight    &    Exposito D    &\citep{Exposito}&    Multi&Time (Multi)&    Opt\\
\midrule
3*3    &5&        11.85    &10.95    &\textbf{9.98} &2.991   &8.78\\
3*4    &5    &    12.20    &11.03    &\textbf{10.33}&  2.932  &    9.03\\
3*5    &5    &    13.18&    11.98&    \textbf{11.60}& 2.812   &10.15\\
3*6    &5    &    14.52&    13.40&    \textbf{13.05}&2.833   &11.28\\
3*7    &5    &    16.77&    15.40&    \textbf{14.80}& 3.058   &12.80\\
3*8    &5    &    17.50&    16.38&    \textbf{15.70}& 3.040   &13.68\\
4*4    &6    &    21.08&    20.10    &\textbf{18.63}&3.243    &15.83\\
4*5    &6    &    25.55&    22.13&    \textbf{21.88}& 3.265   &21.05\\
4*6    &6    &    26.15&    24.20&    \textbf{23.50}& 3.265   &    -\\
4*7    &6    &    30.03&    27.88&    \textbf{27.18}&  3.500   &    -\\
5*5    &7    &    35.03&    31.78&    \textbf{31.48}&3.574     &    -\\
5*6    &7    &    41.30&    38.40&    \textbf{37.30} & 3.996  &-\\
5*7    &7    &    44.50&    41.43&    \textbf{40.73} & 3.872  &-\\
5*8    &7    &    50.35&    47.80&    \textbf{47.25}& 3.873   &    -\\
5*9    &7    &    54.42&    53.73&    \textbf{50.28}&4.088   &    -\\
5*10&    7&    59.05&    58.08&    \textbf{54.38}& 4.271   &    -\\
6*6&    8&    54.80&    51.55&    \textbf{50.23}&  4.047  &    -\\
6*10&    8&    78.45&    77.90&    \textbf{72.40}& 4.941  & -\\
\bottomrule
\end{tabularx}
\end{table}

In Table \ref{table:Multi}, we have also included the optimal results from \citep{Exposito} for smaller instances, and our deterministic implementation of the algorithm from the same article. Our results show that the effect of using randomization, although it always improved the average results, is much larger in case of smaller problem instances. For  larger problem instances the use of a more suitable heuristic is of significantly higher importance. If we observe the results for deterministic algorithms given in Table \ref{table:Fill} we can see that in many cases we outperform the randomized algorithm.

Table \ref{table:Multi} also  shows the execution times (in seconds) for different problem sizes,
where each one of them contains 40 different instances. The approximate calculation time for solving one problem instance
using only one heuristic would be close to 2500 (48*40) times shorter. We wish to emphasize that the implementation
has been done in C\# (which is known for lower speed), and that we did not focus on making highly optimized code. Because of this we believe
that, if necessary, it is possible to develop code of significantly lower calculation time using the same algorithm. The main reason for presenting
these results is to show that the presented method is very suitable for instances of  higher dimensions, due to the good scaling. We can see that the increase of calculation time from the smallest instance (3*3) to the largest(6*10) one is only 1.66 times. The second observation is that the execution time is more dependant of the maximal number of tiers than the number of stacks.

\section{Conclusion}\label{secconcl}

In this paper we have presented a new method for solving the pre-mar\-shalling problem. It can be seen as an improvement to a previously developed heuristic of \citep{Exposito}. We have analyzed different stages of that algorithm, and for each of them we have developed several different heuristics. We have tested and compared the performance of the developed approach on a wide range of problem instances and shown that the newly developed approach outperforms the ones used in the original algorithm in most cases. Our tests have also shown that for the PMP it is very hard to find a universal heuristic that will always give solutions of high quality. We have observed that the performance of the proposed heuristics is highly dependant on the properties of specific problem instances under consideration.

We have used this knowledge to develop a multi-heuristic approach for solving the PMP. The idea of the new method is to exploit the fact that the given greedy heuristic for solving the PMP consists of four stages, and that for each of them several different heuristics exist. We have generated a number of solutions by using a combination of those different heuristics for each stage. In this way only a small group of solutions was generated for which it was expected that they would not have undesirable properties, contrary to the case when simple randomization is used. Our tests have shown that this deterministic algorithm significantly outperforms the original nondeterministic method when the quality of found solutions is observed, with a much lower number of generated solutions.

In the future we plan to develop a more adaptive method for heuristic selection which will provide a higher variation of generating solutions while still avoiding the creation of solutions for which it is expected that they are of lower quality. Moreover, it would be interesting to extend our approach to the problem where each container does not have a specific priority value but some sort of range of priority values, eventually corresponding to prospective changes of priority values, e.g., due to modified ship or truck arrivals.

%
%




%
%

\end{document}